# Optimized Path Planning for Logistics Robots Using Ant Colony Algorithm under Multiple Constraints

Haopeng Zhao
Independent Researcher
Beijing, China
haopeng.zhao1894@gmail.com

Zhichao Ma
Independent Researcher
Shanghai, China
ma.zhi.chao.max@gmail.com

Lipeng Liu
Peking University
Beijing, China
pengpengpku@163.com

Yang Wang
Independent Researcher
Beijing, China
wangyrick102@gmail.com

Zheyu Zhang
Independent Researcher
Beijing, China
zheyuz2980@gmail.com

Hao Liu*
Independent Researcher
Beijing, China
modiy.lu@gmail.com

*Abstract*—With the rapid development of the logistics industry, the path planning of logistics vehicles has become increasingly complex, requiring consideration of multiple constraints such as time windows, task sequencing, and motion smoothness. Traditional path planning methods often struggle to balance these competing demands efficiently. In this paper, we propose a path planning technique based on the Ant Colony Optimization (ACO) algorithm to address these challenges. The proposed method optimizes key performance metrics, including path length, task completion time, turning counts, and motion smoothness, to ensure efficient and practical route planning for logistics vehicles. Experimental results demonstrate that the ACO-based approach outperforms traditional methods in terms of both efficiency and adaptability. This study provides a robust solution for logistics vehicle path planning, offering significant potential for real-world applications in dynamic and constrained environments.

*Keywords- ant colony algorithm; path planning*

## I. INTRODUCTION

The logistics industry plays an important role in the modern economy, and efficient transportation is the cornerstone of its success[1]. However, the path planning of logistics vehicles faces many challenges, especially in urban environments. Logistics vehicles must not only complete tasks within the specified time window[2][3], but also optimize the task sequence to ensure a smooth and energy-efficient route. Traditional path planning methods often fail to fully meet these multi-faceted requirements, resulting in the inability to obtain the optimal solution in real-world scenarios[4][5][6].

In modern urban environments, the path planning problem of logistics vehicles becomes more complicated[7]. In addition to time windows and task sequencing constraints, factors such as traffic congestion, road restrictions, vehicle load, and environmental conditions (such as weather, road conditions, etc.) need to be considered[8]. These factors make it difficult for traditional path planning algorithms to provide effective solutions in dynamic and uncertain environments[9][10]. For example, although the Dijkstra algorithm and the A* algorithm perform well in static environments, they often cannot quickly adapt to real-time traffic changes, resulting in low efficiency[11][12][13]. The ant colony optimization algorithm can effectively find the optimal path in a complex environment by simulating the behavior of ant colonies releasing pheromones during foraging[14][15]. The distributed computing characteristics of ACO enable it to process multiple path selection problems in parallel, so that high-quality solutions can be found in a short time[16][17][18]. In addition, the adaptability of ACO enables it to dynamically adjust the path planning strategy according to environmental changes, making it particularly effective in dealing with dynamic and uncertain problems[19][20][21].

Logistics vehicle path planning usually involves multiple optimization objectives, such as minimizing driving distance, reducing task completion time, reducing energy consumption, and improving path smoothness. These objectives often conflict with each other[22][23]. For example, the shortest path may require more turns, resulting in increased energy consumption and longer driving time. Therefore, finding a balance between these objectives is an important research challenge. This paper adopts the ACO algorithm combined with a multi-objective optimization strategy to consider multiple optimization objectives and find the path with the best overall performance. In order to verify the effectiveness of the proposed method, we designed a series of experiments to simulate the logistics vehicle path planning problem in different urban environments. The experimental results show that the path planning method based on ACO outperforms the traditional algorithm in multiple performance indicators.

## II. METHODOLOGY

### A. Modeling Methods for Complex Constraints in Path Planning

Logistics robots face stringent time window constraints when executing tasks. Each task is associated with a specific time window $[t_{start}, t_{end}]$, within which the robot must arrive at the designated location to complete the task. These time constraints introduce the following challenges in path planning:

Time window conflict: When the time windows of multiple tasks overlap, the robot must intelligently adjust the order of task execution to ensure that all tasks are completed within the specified time. Such conflicts may cause task delays or reduced system efficiency. For example, in a high-density warehouse environment, overlapping time windows can significantly affect the overall throughput of the logistics system.

Path efficiency optimization: While meeting time constraints, the robot must choose the optimal path to minimize the travel distance and time cost. This involves not only path planning in a static environment, but also dynamic obstacle avoidance. Efficient path planning is critical to reducing energy consumption and improving the operational efficiency of the robot fleet.

Dynamic adjustment capability: In the actual operating environment, the robot must respond to unexpected situations in real time and dynamically adjust its path planning strategy. This capability places higher requirements on the robustness and flexibility of the system. For example, in a dynamic warehouse environment, the robot must quickly replan its route to avoid collisions or adapt to new tasks without affecting the overall plan.

**Minimization of Path Length:**

$$f_1 = \min \sum_{i=1}^{n-1} d(p_i, p_{i+1}) \quad (1)$$

where $d(p_i, p_{i+1})$ represents the Euclidean distance between consecutive path points $p_i$ and $p_{i+1}$. Optimizing path length directly impacts the robot's energy consumption and task completion efficiency. Shorter paths reduce travel time and energy usage, which is particularly important in large-scale logistics operations.

**Minimization of Task Completion Time:**

$$f_2 = \min \max_{1 \leq i \leq n} t_i \quad (2)$$

where $t_i$ denotes the completion time of the $i-th$ task. This objective ensures that the robot can complete all tasks in the shortest possible time, thereby improving system throughput. Minimizing the maximum completion time is critical for meeting tight delivery schedules in time-sensitive applications.

**Minimization of Turning Counts:**

$$f_3 = \min \sum_{i=2}^{n-1} \delta(\theta_i) \quad (3)$$

where $\theta_i$ represents the turning angle at the $i-th$ path point, and $\delta(\cdot)$ is an indicator function that takes the value 0, and if the turning angle exceeds a threshold, the value is equal to 1. Reducing turning counts helps improve the robot's motion efficiency and reduces mechanical wear. This is especially important in environments with narrow aisles or limited maneuvering space.

**Optimization of Motion Smoothness:**

$$f_4 = \min \sum_{i=2}^{n-1} |\theta_i - \theta_{i-1}| \quad (4)$$

This objective function ensures smooth motion trajectories, reducing energy loss and mechanical wear caused by sharp turns, while enhancing operational stability. Smooth trajectories also contribute to safer and more predictable robot movements, which is essential in environments shared with human workers.

By establishing a multi-objective optimization model and employing appropriate optimization algorithm, an optimal or near-optimal path planning solution can be found under the constraints, thereby improving the overall efficiency of the logistics system. Additionally, this study considers the robot's dynamic constraints and environmental uncertainties, further enhancing the practicality and robustness of the algorithm. For instance, the algorithm incorporates real-time sensor data to adapt to changing environmental conditions, ensuring reliable performance in complex operational scenarios.

*B. Multi-Constraint Path Planning Based on Ant Colony Algorithm*

The ant colony optimization algorithm is a heuristic optimization algorithm that simulates the foraging behavior of ants. It gradually finds the optimal path by imitating the behavior of ants releasing pheromones in the process of looking for food. This paper combines the ant colony algorithm with the time window constraints and multi-objective optimization constraints in the path planning of logistics robots, and proposes an ant colony algorithm to solve the path planning problem in complex environments.

Based on the traditional ant colony algorithm, this paper introduces time window constraints and multi-objective optimization constraints to improve the pheromone update mechanism and heuristic function design. Pheromone update not only considers the path length, but also combines factors such as task completion time, number of turns, and motion stability. The pheromone update formula is as follows:

$$tau_{ij}(t+1) = (1-\rho) \cdot \tau_{ij}(t) + \Delta \tau_{ij} \quad (5)$$

where $\tau_{ij}(t)$ represents the pheromone concentration on path $(i, j)$, $\rho$ is the pheromone evaporation coefficient, and $\Delta \tau_{ij}$ is the pheromone increment, which is calculated by comprehensively considering path length, task completion time, and motion smoothness.

The heuristic function is used to guide ants in selecting the next path point. The heuristic function is defined as follows:

$$eta_{ij} = \frac{1}{d_{ij}} \cdot \frac{1}{1 + \alpha \cdot \text{penalty}_{ij}} \quad (6)$$

where $d_{ij}$ is the distance of path $(i, j)$, $\text{penalty}_{ij}$ is the turning penalty term, and $\alpha$ is a weight coefficient used to balance path length and turning counts. When selecting the next path point, ants use a probabilistic selection strategy that considers both pheromone concentration and the heuristic function. The path selection probability formula is as follows:

$$p_{ij}^k = \frac{[\tau_{ij}]^\beta [\eta_{ij}]^\gamma}{\sum_{l \in \text{allowed}_k} [\tau_{il}]^\beta [\eta_{il}]^\gamma} \quad (7)$$

where $\beta$ and $\gamma$ are the weight coefficients for pheromone and heuristic function, respectively, and $\text{allowed}_k$ represents the set of nodes that ant $k$ can choose from at the current node.

During the path planning process, the ant needs to check the time window constraints of each task. If the selection of a certain path point will cause the task to be unable to be completed within the specified time, then the path point will be excluded from the optional path points. By dynamically adjusting the path selection strategy, all time window constraints are met. In addition, this paper adopts the weighted summation method to transform the multi-objective optimization problem into a single-objective optimization problem. By adjusting the weight coefficient of each objective function, a balance is achieved between the path length, task completion time, number of turns, and motion smoothness. The objective function is defined as follows:

$$F = w_1 \cdot f_1 + w_2 \cdot f_2 + w_3 \cdot f_3 + w_4 \cdot f_4 \quad (8)$$

where $w_1, w_2, w_3, w_4$ are the weight coefficients for each objective function, satisfying $w_1 + w_2 + w_3 + w_4 = 1$.

The algorithm flow mainly includes four steps: initialization, path construction, pheromone update and iterative optimization. First, the pheromone concentration, heuristic function parameters and starting position of the ants are initialized, and the maximum number of iterations and the number of ants are set. Then each ant gradually builds the path according to the path selection strategy, while checking the time window constraint and multi-objective optimization constraint. After all ants complete the path construction, the pheromone concentration is updated according to the path quality. The higher the path quality, the more pheromone increments. Finally, the path construction and pheromone update process are repeated until the maximum number of iterations is reached or the optimal path that meets the constraints is found.

In order to verify the effectiveness of the ant colony algorithm, experiments were carried out in a simulated warehouse environment. The results show that compared with the traditional ant colony algorithm, the proposed algorithm significantly reduces the path length, the number of turns and the task completion time while meeting the time window constraint and improving the smoothness of the movement. In addition, the robustness of the algorithm in a dynamic environment has been verified, proving that it can effectively handle unexpected situations.

## III. EXPERIMENTS

To validate the effectiveness of the proposed Ant Colony Optimization algorithm for path planning in logistics robots, simulation experiments were designed and compared with classical algorithms in various scenarios. The experimental environment was built on the Gazebo and ROS platforms, simulating real-world warehouse or factory settings with static obstacles (e.g., shelves, walls). The map size was set to 20m×20m with a resolution of 0.1m, and the number of task points ranged from 5 to 20, each associated with a time window $[t_{\text{start}}, t_{\text{end}}]$ randomly distributed between $[5s, 30s]$.

The comparison methods included the Ant Colony Optimization, A*, and Genetic Algorithm (GA), Particle Swarm Optimization(PSO), RRT*, Hybrid A*. The evaluation metrics covered path length, task completion time, turning counts, and motion smoothness. Path length was measured by the total travel distance from the start to the end point; task completion time was the total time for the robot to complete all tasks; turning counts were the total number of turns in the path; and motion smoothness was evaluated by the standard deviation of path curvature, with smaller values indicating smoother paths.

The experiments were conducted in a static environment, and each algorithm was run 50 times for each map complexity level, with the average values taken as the results.

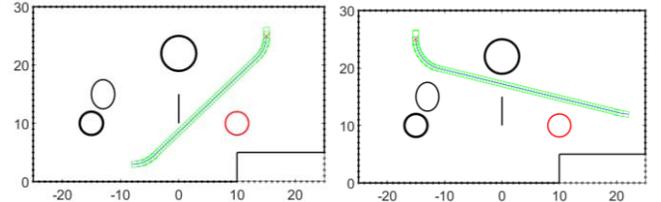

Fig 1. The trajectory of the ACO algorithm.

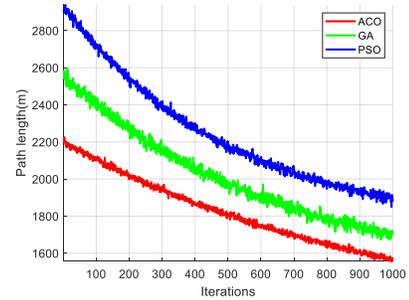

Fig 2. Performance of different algorithms after 1000 iterations.

Fig. 1 shows the performance of ACO algorithm. Since the vehicle turning restriction is added to the optimization target, the vehicle can avoid unnecessary turns. Moreover, when executing different mission objectives, although there are many obstacles between different mission locations, our method can always quickly reach different mission locations without collision. Fig. 2 shows the changes in path length of different intelligent optimization algorithms at 1000 iterations. It can be clearly seen that aco decreases faster and can obtain better results.

TABLE I. PERFORMANCE OF PATH PLANNING

| Method | Length(m) | Time(s) | Truning(rad) | Smoothness(rad) |
|---|---|---|---|---|
| A* | 1349 | 0.12 | 0.4014 | 0.2269 |
| GA | 1455 | 3.78 | 0.3665 | 0.2967 |
| PSO | 1734 | 2.37 | 0.5934 | 0.1571 |
| RRT* | 1627 | 4.61 | 0.4887 | 0.2094 |
| Hybrid A* | 1264 | 0.75 | 0.3316 | 0.192 |
| ACO | 1201 | 0.64 | 0.2793 | 0.1222 |

Table Ⅰ shows that ACO outperforms the comparison algorithms in terms of path length, task completion time, and number of turns. In both simple and complex maps, ACO achieves the shortest path length, the least path planning time, and significantly reduces the number of turns by optimizing the turn penalty. In addition, ACO has the smallest standard deviation of path curvature, indicating the smoothest path.

TABLE II. PERFORMANCE OF TASK COMPLETION

| Method | Task completion(%) | Time(s) |
|---|---|---|
| PSO | 75 | 3.37 |
| GA | 73 | 4.61 |
| ACO | 91 | 3.64 |

Table II shows the task completion of ACO. The results show that, compared with PSO and other methods, ACO is significantly better than other methods in task completion and consumes less time.

## IV. CONCLUSIONS

In this paper, we applied the ACO algorithm to path planning, focusing on optimizing four key performance metrics: Path Length, Task Completion Time, Turning Counts, and Motion Smoothness. The proposed method was extensively evaluated and compared with several existing path planning algorithms. Experimental results demonstrate that the ACO-based approach significantly outperforms the compared methods in terms of efficiency, smoothness, and overall performance. The optimized path planning solution not only reduces travel distance and task completion time but also minimizes unnecessary turns and enhances motion smoothness, making it particularly suitable for real-world applications such as logistics vehicles. Future work will focus on adapting the algorithm for dynamic environments and integrating it into larger-scale logistics systems.